\documentclass[letterpaper, 10 pt, journal, twoside]{IEEEtran} 

\IEEEoverridecommandlockouts                              

\usepackage[dvipsnames]{xcolor}
\definecolor{dark-green}{RGB}{12,80,12}

\usepackage{url}
\usepackage{dsfont}
\usepackage{amsmath}
\usepackage{amssymb}
\usepackage{cuted}
\usepackage{multirow} 
\usepackage{multicol}
\usepackage{flushend}
\usepackage[shortlabels]{enumitem}

\usepackage{tabularx}
\newcolumntype{C}[1]{>{\centering\let\newline\\\arraybackslash\hspace{0pt}}m{#1}} 
\newcolumntype{L}[1]{>{\let\newline\\\arraybackslash\hspace{0pt}}m{#1}} 

\usepackage[mathscr]{euscript}

\usepackage[export]{adjustbox}
\newcolumntype{P}[1]{>{\centering\arraybackslash}p{#1}}
\newcommand{\rot}[1]{\rotatebox[origin=c]{90}{#1}}
\iffalse 
  \newcommand{\todo}[1]{\noindent}
\else
  \newcommand{\todo}[1]{\textcolor{red}{\bf [Todo: #1]}}
  \newcommand\myworries[1]{\textcolor{black}{#1}}
\fi

\usepackage{booktabs}
\usepackage{xspace}
\usepackage{caption}
\usepackage{graphicx} 
\usepackage{color}
\usepackage{siunitx}
\usepackage{subcaption}
\usepackage[hang,flushmargin,symbol]{footmisc}
\renewcommand{\thefootnote}{\fnsymbol{footnote}}
\usepackage{microtype}
\usepackage{rotating}
\newcommand{\secref}[1]{Sec.~\ref{#1}}
\renewcommand{\eqref}[1]{Eq.~(\ref{#1})}
\newcommand{\figref}[1]{Fig.~\ref{#1}}
\newcommand{\tabref}[1]{Tab.~\ref{#1}}

\usepackage{tabularx}
\usepackage{colortbl} 
\newcolumntype{Y}{>{\centering\arraybackslash}X}
\newcolumntype{Z}{>{\raggedleft\arraybackslash}X}

\usepackage{array}
\usepackage{multirow}
\usepackage{pifont}
\usepackage{cite}
\usepackage[export]{adjustbox}
\usepackage[resetlabels]{multibib}
\newcites{New}{References}
\newcommand{\APSref}{mohan2020amodal}
\newcommand{\net}{\mbox{PAPS }}

\DeclareSIUnit{\rad}{rad}

\renewcommand{\baselinestretch}{0.985}

\title{\LARGE \bf
Perceiving the Invisible: Proposal-Free Amodal Panoptic Segmentation
}

\author{Rohit Mohan and Abhinav Valada
\thanks{Department of Computer Science, University of Freiburg, Germany.}%
\thanks{This work was funded by the European Union’s Horizon 2020 research and innovation program under grant agreement No 871449-OpenDR.}
\thanks{Supplementary material available on arXiv.}}

\begin{document}

\maketitle
\thispagestyle{empty}
\pagestyle{empty}

\begin{abstract}
Amodal panoptic segmentation aims to connect the perception of the world to its cognitive understanding. It entails simultaneously predicting the semantic labels of visible scene regions and the entire shape of traffic participant instances, including regions that may be occluded. In this work, we formulate a proposal-free framework that tackles this task as a multi-label and multi-class problem by first assigning the amodal masks to different layers according to their relative occlusion order and then employing amodal instance regression on each layer independently while learning background semantics. We propose the \net architecture that incorporates a shared backbone and an asymmetrical dual-decoder consisting of several modules to facilitate within-scale and cross-scale feature aggregations, bilateral feature propagation between decoders, and integration of global instance-level and local pixel-level occlusion reasoning. Further, we propose the amodal mask refiner that resolves the ambiguity in complex occlusion scenarios by explicitly leveraging the embedding of unoccluded instance masks. Extensive evaluation on the BDD100K-APS and KITTI-360-APS datasets demonstrate that our approach set the new state-of-the-art on both benchmarks. 
\end{abstract}

\section{Introduction}

The ability to perceive the entirety of an object irrespective of partial occlusion is known as amodal perception. This ability enables our perceptual and cognitive understanding of the world~\cite{nanay2018importance}. The recently introduced amodal panoptic segmentation task~\cite{\APSref} seeks to model this ability in robots. The goal of this task is to predict the pixel-wise semantic segmentation labels of the visible amorphous regions of \textit{stuff} classes (e.g., road, vegetation, sky, etc.), and the instance segmentation labels of both the visible and occluded countable object regions of \textit{thing} classes (e.g., cars, trucks, pedestrians, etc.). In this task, each pixel can be assigned more than one class label and instance-ID depending on the visible and occluded regions of objects that it corresponds to, i.e. it allows multi-class and multi-ID predictions. \myworries{Further, for each segment belonging to a \textit{thing} class, the task requires the knowledge of its visible and occluded regions.}

\begin{figure}
    \centering
     \includegraphics[width=0.5\textwidth]{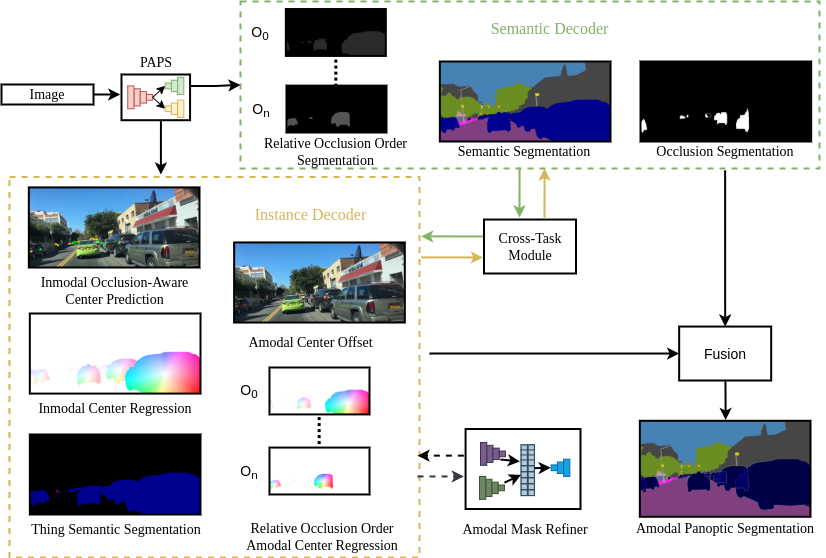}
    \caption{ Overview of our proposed \net architecture for amodal panoptic segmentation. Our model predicts multiple outputs from both the semantic and instance decoder. We then fuse the instance-agnostic semantic labels and foreground masks obtained from the segmentation heads with class-agnostic amodal instances that are obtained from the rest of the heads by grouping and majority voting to yield the final amodal panoptic segmentation output.}
    \label{fig:paper-teaser}
   \vspace{-0.4cm}
\end{figure}

The existing amodal panoptic segmentation approach~\cite{\APSref} and baselines~\cite{\APSref} follow the proposal-based architectural topology. Proposal-based methods tend to generate overlapping inmodal instance masks as well as multiple semantic predictions for the same pixel, one originating from the instance head and the other from the semantic head, which gives rise to a conflict when fusing the task-specific predictions. This problem is typically tackled using cumbersome heuristics for fusion, requiring multiple sequential processing steps in the pipeline which also tends to favor the amodal instance segmentation branch. On the other hand, proposal-free methods have been more effective in addressing this problem in the closely related panoptic segmentation task~\cite{sofiiuk2019adaptis, Gao_2019_ICCV, cheng2020panoptic} by directly predicting non-overlapping segments. In this work, we aim to alleviate this problem by introducing the first proposal-free framework called Proposal-free Amodal Panoptic Segmentation (PAPS) architecture to address the task of amodal panoptic segmentation. Importantly, to facilitate multi-class and multi-ID predictions, our \net decomposes the amodal masks of objects in a given scene into several layers based on their relative occlusion ordering in addition to conventional instance center regression for visible object regions of the scene referred to as inmodal instance center regression. Hence, the network can focus on learning the non-overlapping segments present within each layer. \figref{fig:paper-teaser} illustrates an overview of our approach.

Further, amodal panoptic segmentation approaches tend to predict the amodal masks of \textit{thing} class objects by leveraging occlusion features that are conditioned on features of the visible regions. Although it is effective when objects are only partially occluded, it fails in the presence of heavy occlusion as the area of the visible region is reduced. Motivated by humans whose amodal perception is not only based on visible and occlusion cues but also their experience in the world, we propose the amodal mask refiner module to model this capability using explicit memory. This module first predicts an embedding that represents the unoccluded object regions and correlates it with the amodal features generated using either a proposal-free or proposal-based method to complement the lack of visually conditioned occlusion features. We also demonstrate that our amodal mask refiner can be readily incorporated into a variety of existing architectures to improve performance.

An interesting aspect of proposal-free methods is that the two sub-tasks, namely, semantic segmentation and instance center regression, are complementary in nature. We leverage this to our benefit and propose a novel cross-task module to bilaterally propagate complementary features between the two sub-tasks decoders for their mutual benefit. Moreover, as rich multi-scale features are important for reliable instance center prediction, we propose the context extractor module that enables within-scale and cross-scale feature aggregation. Finally, to exploit informative occlusion features that play a major role in the amodal mask segmentation quality~\cite{\APSref, qi2019amodal}, we incorporate occlusion-aware heads in our \net architecture to capture local pixel-wise and global instance-level occlusion information. We present extensive quantitative and qualitative evaluations of \net on the challenging BDD100K-APS and KITTI-360-APS datasets, which shows that it achieves state-of-the-art performance. Additionally, we present comprehensive ablation studies to demonstrate the efficacy of our proposed architectural components and we make the models publicly available at \url{http://amodal-panoptic.cs.uni-freiburg.de}. 

\section{Related Work}
\label{sec:relatedWork}

Although the amodal panoptic segmentation task~\cite{\APSref} is relatively new, the inmodal variant called panoptic segmentation has been extensively studied. We first briefly discuss the methods for panoptic segmentation followed by amodal panoptic segmentation.   

{\parskip=5pt
\noindent\textit{Panoptic Segmentation}: We can categorize existing methods into top-down and bottom-up approaches. Top-down approaches~\cite{porzi2019seamless, gosala2022bird, mohan2020efficientps,sirohi2021efficientlps} follow the topology of employing task-specific heads, where the instance segmentation head predicts bounding boxes of objects and its corresponding mask, while the semantic segmentation head outputs the class-wise dense semantic predictions. Subsequently, the outputs of these heads are fused by heuristic-based fusion modules~\cite{xiong2019upsnet, mohan2020efficientps}. On the other hand, bottom-up panoptic segmentation methods~\cite{Gao_2019_ICCV, cheng2020panoptic} first perform semantic segmentation, followed by employing different techniques to group~\cite{leibe2004combined,uhrig2018box2pix,valverde2021there} \textit{thing} pixels to obtain instance segmentation. In this work, we follow the aforementioned schema with instance center regression to obtain the panoptic variant of our proposed architecture. Our proposed network modules enrich multi-scale features by enabling feature aggregation from both within-scales and cross-scales. Additionally, our cross-task module facilitates the propagation of complementary features between the different decoders for their mutual benefit.}

{\parskip=5pt
\noindent\textit{Amodal Panoptic Segmentation}: Mohan~\textit{et~al.}~\cite{\APSref} propose several baselines for amodal panoptic segmentation by replacing the instance segmentation head of EfficientPS~\cite{mohan2020efficientps}, a top-down panoptic segmentation network, with several existing amodal instance segmentation approaches. EfficientPS employs a shared backbone comprising of an encoder and the 2-way feature pyramid in conjunction with a Mask R-CNN based instance head and a semantic segmentation head, whose outputs are fused to yield the panoptic segmentation prediction. The simple baseline, Amodal-EfficientPS~\cite{\APSref}, extends EfficientPS with an additional amodal mask head and relies implicitly on the network to capture the relationship between the occluder and occludee. ORCNN~\cite{follmann2019learning} further extends it with an invisible mask prediction head to explicitly learn the feature propagation from inmodal mask to amodal mask. Subsequently, ASN~\cite{qi2019amodal} employs an occlusion classification branch to model global features and uses a multi-level coding block to propagate these features to the individual inmodal and amodal mask prediction heads. More recently, Shape Prior~\cite{yuting2021amodal} focuses on leveraging shape priors using a memory codebook with an autoencoder to further refine the initial amodal mask predictions. Alternatively, VQ-VAE~\cite{jang2020learning} utilizes shape priors through discrete shape codes by training a vector quantized variational autoencoder. BCNet~\cite{Ke_2021_CVPR} seeks to decouple occluding and occluded object instances boundaries by employing two overlapping GCN layers to detect the occluding objects and partially occluded object instances. The most recent, APSNet~\cite{\APSref} which is the current state-of-the-art top-down approach focuses on explicitly modeling the complex relationships between the occluders and \myworries{occludees}. To do so, APSNet employs three mask heads that specialize in segmenting visible, occluder, and occlusion regions. It then uses a transformation block with spatio-channel attention for capturing the underlying inherent relationship between the three heads before computing the final outputs. In this work, we present the first bottom-up approach that learns the complex relationship between the occluder and occludee by focusing on learning the relative occlusion ordering of objects. We also employ an occlusion-aware head to explicitly incorporate occlusion information and an amodal mask refiner that aims to mimic the ability of humans by leveraging prior knowledge on the physical structure of objects for amodal perception.}



\section{Methodology}
\label{sec:method}

\begin{figure*}
    \centering
    \begin{subfigure}[b]{\linewidth}
        \centering
        \includegraphics[width=\textwidth]{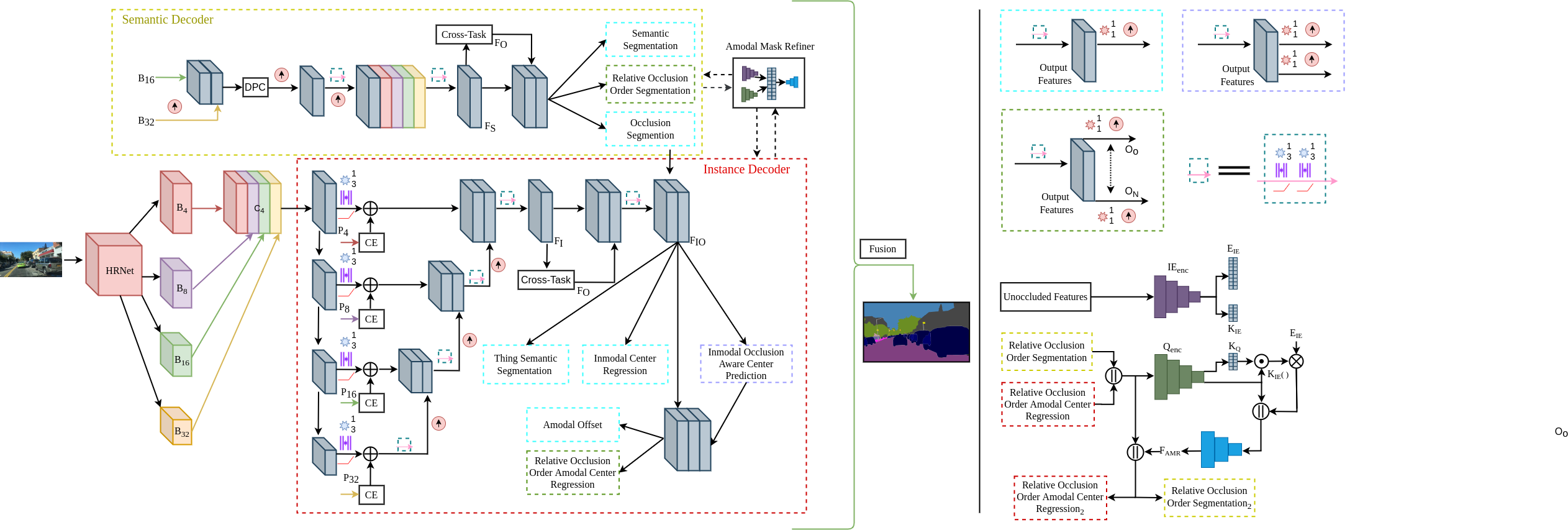}
        \label{APSNEt_arch}
    \end{subfigure}
    \begin{subfigure}[b]{0.8\linewidth}
        \centering
        \includegraphics[width=\textwidth]{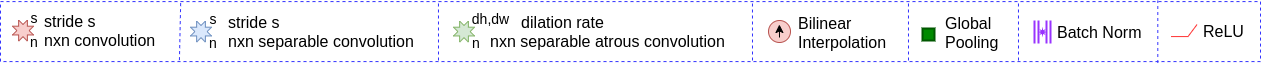}
    \end{subfigure}
    \caption{\myworries{Illustration of our proposed \mbox{PAPS} architecture consisting of a shared backbone and asymmetric dual-decoder followed by a fusion module that fuses the outputs of the multiple heads of the decoder to yield the amodal panoptic segmentation output. The semantic decoder (yellow-green) and the instance decoder (dark-red) boxes show the topologies of the dual-decoder employed in our architecture. The black-box shows the architecture of our proposed context extractor module. The amodal mask refiner module exploits features from both the decoders to improve amodal masks with embedding correlation.}}
    \label{fig:network}
    \vspace{-0.3cm}
\end{figure*}

In this section, we first describe our \net architecture and then detail each of its constituting components. \figref{fig:network} illustrates the network which follows the bottom-up topology. It consists of a shared backbone followed by semantic segmentation and amodal instance segmentation decoders. The outputs of the decoders are then fused during inference to yield the amodal panoptic segmentation predictions. \net incorporates several novel network modules to effectively capture multi-scale features from within-layers and cross-layers, to enable bilateral feature propagation between the task-specific decoders and exploit local and global occlusion information. Further, it incorporates our amodal mask refiner that embeds unoccluded inmodal instance masks to refine the amodal features. 

\subsection{PAPS Architecture}
\label{sec:architecture}

\subsubsection{Backbone}

The backbone is built upon HRNet~\cite{wang2020deep} which specializes in preserving high-resolution information throughout the network. It has four parallel outputs with a scale of $\times4$, $\times8$, $\times16$ and $\times32$ downsampled with respect to the input, namely, B\textsubscript{4}, B\textsubscript{8}, B\textsubscript{16}, and B\textsubscript{32}, as shown in \figref{fig:network}. We then upsample the feature maps to $\times4$ and concatenate the representations of all the resolutions resulting in C\textsubscript{4}, followed by reducing the channels to $256$ with a $1\times1$ convolution. Lastly, we aggregate multi-scale features by downsampling high-resolution representations to multiple levels and process each level with a $3\times3$ convolution layer (P\textsubscript{4}, P\textsubscript{8}, P\textsubscript{16}, P\textsubscript{32}).

\subsubsection{Context Extractor}

The multi-scale representations from the backbone are computed over all four scales which we refer to as cross-scale features. The way these cross-scale features are computed (concatenation, reduction, and downsampling) leads to a limited exploration for multi-scale features at a given individual scale resolution. Since rich multi-scale representations are crucial for the instance decoder's performance, we seek to enhance the cross-scale features with within-scale contextual features. To do so, we design a lightweight module called the context extractor which is based on the concept of spatial pyramid pooling and is known for efficiently capturing multi-scale contexts from a fixed resolution. We use the context extractor module at each scale (B\textsubscript{4}, B\textsubscript{8}, B\textsubscript{16}, B\textsubscript{32}) , and add its output to P\textsubscript{4}, P\textsubscript{8}, P\textsubscript{16}, and P\textsubscript{32}, respectively. The proposed context extractor module shown in \myworries{Fig. S.1 in the supplementary material}, employs two $1\times1$ convolutions, two $3\times3$ depth-wise atrous separable convolutions with a dilation rate of $(1,6)$ and $(3,1)$, respectively, and a global pooling layer. The output of this module consists of $256$ channels, where $128$ channels are contributed by the $1\times1$ convolution and four $32$ channels come from each of the two $3\times3$ depth-wise atrous separable convolutions and its globally pooled outputs. 
We evaluate the benefits of the aforementioned module in the ablation study presented in \secref{subsubsec:panop}.\looseness=-1

\subsubsection{Cross-Task Module}

The sub-tasks, semantic segmentation and amodal instance center regression, are both distinct recognition problems and yet closely related. The intermediate feature representations of each task-specific decoder can capture complementary features that can assist the other decoder to improve its performance. We propose the cross-task module to enable bilateral feature propagation between the decoders to mutually benefit each other. Given feature inputs $F_{I}$ and $F_{S}$ from the two decoders, we fuse them adaptively by employing cross-attention followed by self-attention as
\begin{align}
    \label{eq:cross}
    F_{R} &= (1-g_1(F_{S}))\cdot F_{I} + (1-g_2(F_{I}))\cdot F_{S},\\
    \label{eq:cross2}
    F_{O} &= g_3(F_{R})\cdot F_{R},
\end{align}
where $g_1(\cdot)$, $g_2(\cdot)$, and $g_3(\cdot)$ are functions to compute feature confidence score of $F_{S}$ and $F_{I}$. These functions consist of a global pooling layer, followed reducing the channels from $256$ to $64$ using a $1\times1$ convolution. Subsequently, we employ another $1\times1$ convolution with $256$ output channels to remap from the lower dimension to a higher dimension and apply a sigmoid activation to obtain the feature confidence scores. $F_{O}$ is the output of the cross-task module. The cross-attention mechanism in this module enables $F_{I}$ and $F_{S}$ to adaptively complement each other, whereas the following self-attention mechanism enables enhancing the highly discriminative complementary features. The ablation study presented in \secref{subsubsec:panop} shows the influence in performance due to this module.

\subsubsection{Semantic Decoder}
\label{subsubsec:semantic}

The semantic decoder takes B\textsubscript{32}, B\textsubscript{16}, C\textsubscript{4} feature maps and the output of cross-task module as its input. First, the B\textsubscript{32} feature maps are upsampled ($\times16$) and concatenated with B\textsubscript{16} and are fed to the dense prediction cell (DPC)~\cite{chen2018searching}. The output of DPC is then upsampled ($\times8$) and passed through two sequential $3\times3$ depth-wise separable convolutions. Subsequently, we again upsample ($\times4$) and concatenate it with C\textsubscript{4}. We then employ two sequential $3\times3$ depth-wise separable convolutions and feed the output ($F_{S}$) to the cross-task module. Further, we concatenate $F_{S}$ with the output of the cross-task module ($F_{O}$) and feed it to the multiple heads in the semantic decoder.

\begin{figure}
    \footnotesize
    \centering
    \includegraphics[width=0.48\textwidth]{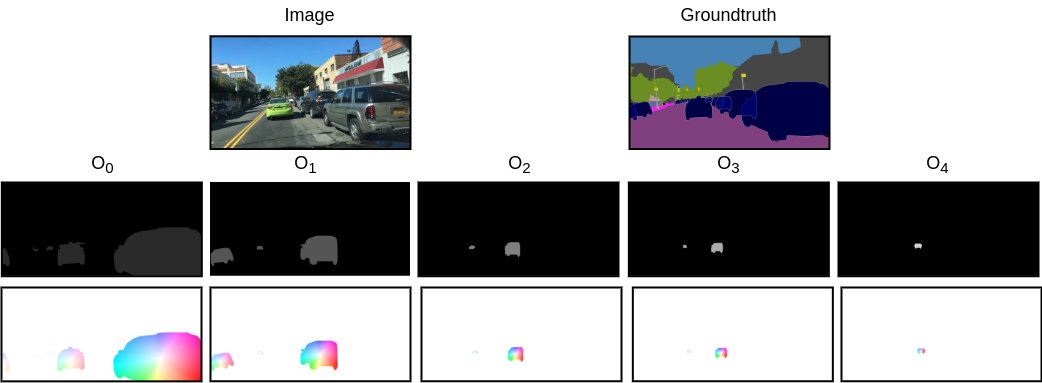}
    \caption{Groundtruth examples for relative occlusion order segmentation (top-row) and instance center regression (bottom-row) consisting of layer from O\textsubscript{0} to O\textsubscript{5}. Best viewed at $\times4$ zoom.}
    \label{fig:roo_eg}
    \vspace{-0.3cm}
\end{figure}

We employ three heads, namely, relative occlusion order segmentation \myworries{(${L}_{roo}$)}, semantic segmentation \myworries{(${L}_{ss}$)}, and occlusion segmentation \myworries{(${L}_{os}$)}, towards the end of our semantic decoder. The relative occlusion order segmentation head predicts foreground mask segmentation for $O_{N}$ layers. The masks of each layer are defined as follows: All unoccluded class-agnostic \textit{thing} object masks belong to layer $0$ (O\textsubscript{0}). Next, layer $1$ (O\textsubscript{1}) comprises amodal masks of any occluded object that are occluded by layer $0$ objects but not occluded by any other occluded object. Next, layer $2$ (O\textsubscript{2}) consists of amodal masks of any occluded object, not in the previous layers that are occluded by layer $1$ objects but not occluded by any other occluded objects that are not part of previous layers and so on. \figref{fig:roo_eg} illustrates the separation of \textit{thing} amodal object segments into relative occlusion ordering layers. This separation ensures each \textit{thing} amodal object segment belongs to a unique layer without any overlaps within that layer. We use the binary cross-entropy loss (${L}_{roo}$) to train this head. Next, the semantic segmentation head predicts semantic segmentation of both \textit{stuff} and \textit{thing} classes, and we employ the weighted bootstrapped cross-entropy loss~\cite{yang2019deeperlab} (${L}_{ss}$) for training. Lastly, the occlusion segmentation head predicts whether a pixel is occluded in the context of \textit{thing} objects and we use the binary cross-entropy loss (${L}_{occ}$) for training. The overall semantic decoder loss is given as\looseness=-1
\begin{equation}
{L_{sem}} = {L}_{ss}+ \myworries{{L}_{os}} + {L}_{roo}.
\end{equation}
The predictions from all the heads of the semantic decoder are used in the fusion module to obtain the final amodal panoptic segmentation prediction.

\subsubsection{Instance Decoder}
\label{subsubsec:instance}

The instance decoder employs a context encoder at each scale (B\textsubscript{32}, B\textsubscript{16}, B\textsubscript{8}, B\textsubscript{4}) and adds the resulting feature maps to P\textsubscript{32}, P\textsubscript{16}, P\textsubscript{8}, and P\textsubscript{4}, respectively. Then, beginning from ($\times32$), the decoder repeatedly uses a processing block consisting of two sequential $3\times3$ depth-wise separable convolutions, upsamples it to the next scale ($\times16$), and concatenates with the existing features of the next scale until $\times4$ feature resolution is obtained ($F_{I}$). The $F_{I}$ is then fed to the cross-task module. The cross-task output $F_O$ is concatenated with $F_{I}$ and is processed by two sequential $3\times3$ depth-wise separable convolutions. Subsequently, the features from the occlusion segmentation head of the semantic decoder are concatenated to incorporate explicit pixel-wise local occlusion information referred to as $F_{IO}$ features.

The instance decoder employs five prediction heads. The inmodal occlusion-aware center prediction head consists of two prediction branches, one for predicting the center of mass heatmap of inmodal \textit{thing} object instances \myworries{(${L}_{icp}$)} and the other for predicting whether the heatmap is occluded \myworries{(${L}_{ico}$)}. For the former, we use the Mean Squared Error (MSE) loss (${L}_{icp}$) to minimize the distance between the 2D Gaussian encoded groundtruth heatmaps and the predicted heatmaps, for training. For the latter, we use binary cross-entropy loss \myworries{(${L}_{ico}$)} for training. Following, the \textit{thing} semantic segmentation \myworries{(${L}_{tss}$)} head predicts N\textsubscript{\textit{thing}}+1 classes, where N\textsubscript{\textit{thing}} is the total number of \textit{thing} semantic classes and the '$+1$' class predicts all \textit{stuff} classes as a single class. This head is trained with the weighted bootstrapped cross-entropy loss~\cite{yang2019deeperlab} (${L}_{tss}$). Next, the inmodal center regression \myworries{(${L}_{icr}$)} head predicts the offset from each pixel location belonging to \textit{thing} classes to its corresponding inmodal object instance mass center. We use the $L_1$ loss for training this head (${L}_{icr}$). All the aforementioned heads take $F_{IO}$ features as input. 

The remaining heads of the instance decoder are referred to as the amodal center offset \myworries{(${L}_{aco} $)} and relative occlusion order amodal center regression \myworries{(${L}_{rooacr} $)}. The amodal center offset head predicts the offset from each inmodal object instance center to its corresponding amodal object instance center. Whereas, the relative occlusion ordering amodal center regression head, for each relative occlusion ordering layer, predicts the offset from each pixel location belonging to \textit{thing} classes of the layer to its corresponding amodal object instance mass center. Here, the layers of relative occlusion ordering are defined similarly as in the semantic decoder. Further, we concatenate $F_{IO}$ with features of inmodal occlusion-aware center prediction head to incorporate object-level global occlusion features before feeding it to the aforementioned heads. Finally, we use $L_1$ loss to train both the heads (${L}_{aco} $, ${L}_{rooacr}$). The overall loss for the instance decoder is
\begin{equation}
{L_{inst}} = {L}_{tss}+ \myworries{{L}_{ico}} + \alpha{L}_{icp} + \beta({L}_{icr} + {L}_{aco} + {L}_{rooacr}),
\end{equation}
where the loss weights $\alpha=200$ and $\beta=0.01$.

Note that we learn amodal center offset instead of the amodal center itself to have a common instance-ID that encapsulates both the amodal and inmodal masks.

\subsubsection{Amodal Mask Refiner}
\label{sec:memory}

We propose the amodal mask refiner module to model the ability of humans to leverage priors on complete physical structures of objects for amodal perception, in addition to visually conditioned occlusion cues. This module builds an embedding that embeds the features of the unoccluded object mask and correlates them with the generated amodal features to complement the lack of visually conditioned occlusion features. The amodal mask refiner \myworries{shown in \figref{fig:network}} consists of two encoders, unoccluded feature embeddings, and a decoder. We employ the RegNet~\cite{radosavovic2020designing} topology with its first and last stages removed as the two encoders with feature encoding resolution of $\times16$ downsampled with respect to the input. The two encoders are an inmodal embedding encoder (IE\textsubscript{enc} $\in \mathbb{R}^{(H/16)\times(W/16)\times C}$) that encodes unoccluded objects features and a query encoder (Q\textsubscript{enc} $\in \mathbb{R}^{(H/16)\times(W/16)\times C}$) that encodes the amodal features, where $H$ and $W$ are the height and width of the input image and $C$ is the feature dimension which is set to $64$. Subsequently, an embedding matrix E\textsubscript{IE} $\in  \mathbb{R}^{N\times D}$ embeds the IE\textsubscript{enc} encoding to create the embedding of unoccluded object masks. Further, to extract the mask embedding information from E\textsubscript{IE}, we compute two key matrices, namely, K\textsubscript{IE} $\in  \mathbb{R}^{N\times D}$ matrix and K\textsubscript{Q} $\in \mathbb{R}^{1\times D}$ matrix, from IE\textsubscript{enc} and Q\textsubscript{enc} encodings, respectively. Here, $N=128$ and $D=[(H/16)\times(W/16)\times C]$.


Next, we compute the inner product of K\textsubscript{IE} and K\textsubscript{Q} followed by a softmax and take the inner product of the resulting probability and E\textsubscript{IE}. We then rearrange this output into $(H/16)\times(W/16)\times C$ shape and concatenate it with Q\textsubscript{enc} and feed it to the decoder. The decoder employs repeated blocks of two $3\times3$ depth-wise separable convolutions, followed by a bilinear interpolation to upsample by a factor of 2 until the upsampled output resolution is $\times4$ downsampled with respect to the input. We refer to this output as F\textsubscript{AMR}. The resulting features enrich the amodal features of occluded objects with similar unoccluded object features, thereby enabling our network to predict more accurate amodal masks. 

The amodal mask refiner takes two inputs, namely, the amodal features and the features of the unoccluded objects. The input amodal features are obtained by concatenating the output features (\figref{fig:network}) of relative occlusion ordering heads of the semantic and instance decoders. To compute the features of the unoccluded object, we first perform instance grouping using predictions of the inmodal occlusion-aware, inmodal center regression, and \text{thing} semantic segmentation heads to obtain the inmodal instance masks. We then discard all the occluded inmodal instances to generate an unoccluded instance mask. Next, we multiply the aforementioned mask with the output of the second layer of the inmodal center regression head to compute the final unoccluded object features. Finally, the amodal mask refiner outputs F\textsubscript{AMR} which is then concatenated with the amodal features. We employ two similar heads as relative occlusion ordering amodal center regression and segmentation that takes the aforementioned concatenated features as input. We use the same loss functions and loss weights for training the heads as described in~\secref{subsubsec:instance}.

\subsubsection{Inference}
\label{subsubsec:inference}

We perform a series of steps during inference to merge the outputs of the semantic and instance decoders to yield the final amodal panoptic segmentation. We begin with computing the semantic segmentation prediction and the \textit{thing} foreground mask. To do so, we duplicate the void class logit of the \textit{thing} semantic segmentation head logits $N_{stuff}$-times, such that its number of channels transforms from $1+N_{thing}$ to $N_{stuff}+N_{thing}$. We then add it to the logits of the semantic segmentation head and employ a softmax followed by an argmax function to obtain the final semantic segmentation prediction. Subsequently, we assign $0$ to all the \textit{stuff} classes and $1$ to all the \textit{thing} classes to obtain the \textit{thing} foreground mask. Next, we obtain the inmodal center point predictions by employing a keypoint-based non-maximum suppression~\cite{cheng2020panoptic} and confidence thresholding ($0.1$) to filter out the low confidence predictions while keeping only the top-k ($200$) highest confidence scores on the heatmap prediction of inmodal occlusion-aware center prediction head. We then obtain the amodal center points predictions by applying the corresponding offsets from the amodal instance head to the inmodal center point predictions. We obtain the class-agnostic instance-IDs and the inmodal instance mask using simple instance grouping~\cite{cheng2020panoptic} with the inmodal center prediction and the \textit{thing} foreground mask. Further, we compute semantic labels for each instance-ID by the majority vote of the corresponding predicted semantic labels with its inmodal instance masks.

Now, for each instance-ID, we have its semantic label, inmodal mask, and the amodal center prediction. We compute the relative occlusion order segmentation masks for each layer by applying a threshold of $0.5$ on the outputs of the relative occlusion ordering segmentation head connected to the amodal mask refiner. We then assign the instance-ID to its corresponding relative occlusion ordering layer by checking if the corresponding amodal center lies within the segmentation mask of the layer in question. Finally, we again use the simple instance grouping at each of the relative occlusion ordering layers. For all instance-IDs belonging to a layer, we apply the instance grouping using its amodal instance center and regression along with the corresponding segmentation mask to compute the amodal mask. In the end, for each \textit{thing} object, we have its unique instance-ID, semantic label, inmodal, and amodal mask along with \textit{stuff} class semantic predictions from the semantic segmentation prediction. \myworries{We obtain the visible attribute of the amodal mask directly from the inmodal mask and obtain the occluded attributes of the amodal mask by removing the inmodal mask segment from the amodal mask.}

\section{Experimental Evaluation}
\label{sec:experiments}

\begin{table*}
\footnotesize 
\centering
\caption{Comparison of amodal panoptic segmentation benchmarking results on the KITTI-360-APS and BDD100K-APS validation set. Subscripts $S$ and $T$ refer to \textit{stuff} and \textit{thing} classes respectively. All scores are in [\%].}
\begin{tabular}{p{3cm}|p{0.6cm}p{0.6cm}p{0.6cm}p{0.6cm}p{0.6cm}p{0.6cm}|p{0.6cm}p{0.6cm}p{0.6cm}p{0.6cm}p{0.6cm}p{0.6cm}}
\toprule
Model &  \multicolumn{6}{c|}{KITTI-360-APS} & \multicolumn{6}{c}{BDD100K-APS}\\
\cmidrule{2-13}
 &  APQ  & APC & APQ$_S$ &APQ$_T$ & APC$_S$&  APC$_T$  & APQ  & APC & APQ$_S$ &APQ$_T$ & APC$_S$&  APC$_T$ \\
\midrule
Amodal-EfficientPS & $41.1$ & $57.6$ & $46.2$ & $33.1$ & $58.1$ & $56.6$ &  $44.9$ & $46.2$ & $54.9$ & $29.9$ & $64.7$ & $41.4$ \\
ORCNN~\cite{follmann2019learning}  & $41.1$ & $57.5$ & $46.2$ & $33.1$ & $58.1$ & $56.6$ & $44.9$ & $46.2$ & $54.9$ & $29.9$ & $64.7$ & $41.5$  \\
BCNet~\cite{Ke_2021_CVPR} & $41.6$ & $57.9$ & $46.2$ & $34.4$ & $58.1$ & $57.6$  & $45.2$ & $46.4$ & $55.0$ & $30.7$ & $64.7$ & $42.1$ \\
VQ-VAE~\cite{jang2020learning}  & $41.7$ & $58.0$  & $46.2$ & $34.6$ & $58.1$ & $57.8$ & $45.3$ & $46.5$ & $54.9$ & $30.8$ & $64.7$ & $42.2$  \\
Shape Prior~\cite{yuting2021amodal}  & $41.8$ & $58.2$  & $46.2$ & $35.0$ & $58.1$ & $58.2$& $45.4$ & $46.6$ & $55.0$ & $31.0$ & $64.8$ & $42.6$   \\
ASN~\cite{qi2019amodal} & $41.9$ & $58.2$ & $46.2$ & $35.2$ & $58.1$ & $58.3$ & $45.5$ & $46.6$ & $55.0$ & $31.2$ & $64.8$ & $42.7$   \\
\mbox{APSNet}~\cite{\APSref} &  $42.9$ & $59.0$  & $46.7$ & $36.9$  & $58.5$ & $59.9$  & $46.3$ & $47.3$  & $55.4$ & $32.8$  & $65.1$ & $44.5$  \\
\midrule
\net (Ours) &  $\mathbf{44.6}$ & $\mathbf{61.4}$  & $\mathbf{47.5}$ & $\mathbf{40.1}$  & $\mathbf{59.2}$ & $\mathbf{64.7}$ & $\mathbf{48.7}$ & $\mathbf{50.4}$  & $\mathbf{56.5}$ & $\mathbf{37.1}$  & $\mathbf{66.4}$ & $\mathbf{51.6}$ \\
\bottomrule
\end{tabular}
\label{tab:kittiEvaluation}
\vspace{-0.3cm}
\end{table*}

In this section, we describe the datasets that we benchmark on in~\secref{subsec:datasets} and the training protocol in~\secref{subsec:training}. We then present extensive benchmarking results in~\secref{subsec:benchmarking}, followed by a detailed ablation study on the architectural components in~\secref{subsec:ablation} and qualitative comparisons in \secref{subsec:qualitative}. We use the standard Amodal Panoptic Quality (APQ) and Amodal Parsing Coverage (APC) metrics~\cite{mohan2020amodal} to quantify the performance.\looseness=-1

\subsection{Datasets}
\label{subsec:datasets}

\textit{KITTI-360-APS}~\cite{\APSref} provides amodal panoptic annotations for the KITTI-360~\cite{Liao2021ARXIV} dataset. It consists of $9$ sequences of urban street scenes with annotations for $61{,}168$ images. The sequence numbered $10$ of the dataset is treated as the validation set. This dataset comprises $7$ \textit{thing} classes, namely, car, pedestrians, cyclists, two-wheeler, van, truck, and other vehicles. Further, the dataset consists of $10$ \textit{stuff} classes. These stuff classes are road, sidewalk, building, wall, fence, pole, traffic sign, vegetation, terrain, and sky. 

\textit{BDD100K-APS}~\cite{\APSref} extends the BDD100K~\cite{yu2020bdd100k} dataset with amodal panoptic annotations for $15$ of its sequences consisting of $202$ images per sequence. The training and validation set consists of $12$  and $3$ sequences, respectively. Pedestrian, car, truck, rider, bicycle, and bus are the $6$ \textit{thing} classes. Whereas, road, sidewalk, building, fence, pole, traffic sign, fence, terrain, vegetation, and sky are the $10$ \textit{stuff} classes

\subsection{Training Protocol}
\label{subsec:training}

All our models are trained using the PyTorch library on $8$ NVIDIA TITAN RTX GPUs with a batch size of 8. We train our network in two stages, with a crop resolution of $376\times1408$~pixels and $448\times1280$~pixels for the KITTI-360-APS and BDD100K-APS datasets, respectively. For each stage, we use the Adam optimizer with a poly learning rate schedule, where the initial learning rate is set to $0.001$. We train our model for $300$K iterations for the KITTI-360-APS dataset and $70$K iterations for the BDD100K-APS dataset, while using random scale data augmentation within the range of $[0.5,2.0]$ with flipping for each stage. \myworries{We use $N=8$ for relative occlusion order layers}. We first train the model without the amodal mask refiner, followed by freezing the weights of the architectural components from the previous stage and train only the amodal mask refiner.

\subsection{Benchmarking Results}
\label{subsec:benchmarking}

In this section, we present results comparing the performance of our proposed \net architecture against current state-of-the-art amodal panoptic segmentation approaches. We report the APQ and APC metrics of the existing state-of-the-art methods directly from the published manuscript~\cite{\APSref}. \tabref{tab:kittiEvaluation} presents the benchmarking results on both datasets.

We observe that our proposed \net architecture achieves the highest APQ and APC scores compared to APSNet and other baselines on both datasets. The improvement of $1.7\%\text{-}2.7\%$ in both the metrics can be attributed to the various proposed components of our architecture. For \textit{stuff} segmentation, the complementary features from the cross-task module aid in better distinguishing \textit{stuff} and \textit{thing} classes, while the high resolution features with the long-range contextual features help in finer segmentation of the boundaries. Consequently, we observe an improvement of $0.7\%\text{-}1.3\%$ in the \textit{stuff} components of the metrics for both datasets. The \textit{thing} components of the metrics achieve an improvement of $3.2\%\text{-}7.1\%$ which can be attributed to the synergy of several factors. The context extractor and the cross-task modules provide richer multi-scale representations along with complementary semantic decoder features. This enables reliable segmentation of far-away small-scale instances. Further, the incorporation of local and object-level global occlusion information from the instance and semantic decoder heads enables explicit amodal reasoning capabilities. We also believe that the relative occlusion ordering layers force the network to capture the complex underlying relationship of objects to one another in the context of occlusions. Lastly, the amodal mask refiner module with its transformation of amodal features with unoccluded object mask embeddings improves the quality of large occlusion area segmentation as observed from the higher improvement in APC than the APQ metric. Overall, \net establishes the new state-of-the-art on both the amodal panoptic segmentation benchmarks.  


\subsection{Ablation Study}
\label{subsec:ablation}

In this section, we first study the improvement due to the various architectural components that we propose in our \net and study the generalization ability of the amodal mask refiner by incorporating it in various proposal-based methods. We then evaluate the performance of \net for panoptic segmentation and amodal instance segmentation tasks.

\subsubsection{Detailed Study on the \net Architecture}
\label{subsubsec:panop}


\begin{table}
\footnotesize 
\centering
\caption{Evaluation of various architectural components of our proposed \net model. The performance is shown for the models trained on the BDD100K-APS dataset and evaluated on the validation set. Subscripts $S$ and $T$ refer to \textit{stuff} and \textit{thing} classes respectively. All scores are in [\%].}
\label{tab:instanceHeadEvaluation}
\begin{tabular}{p{1.7cm}|p{0.6cm}p{0.6cm}|p{0.6cm}p{0.6cm}|p{0.6cm}p{0.6cm}}
\toprule
Model & APQ & APC &  APQ$_S$ &APQ$_T$ & APC$_S$ & APC$_T$\\
\midrule
M1  &$45.6$ &  $46.9$ &$55.8$ &$30.4$ & $65.7$ & $42.2$   \\
M2  &$45.9$ &   $47.1$ &$55.8$ &$31.0$ &$65.7$ & $42.7$   \\
M3  & $46.1$ &   $47.2$ &$55.9$ &$31.3$ &$65.8$ & $42.9$  \\
M4  &$46.3$ &   $47.3$ &$55.9$ &$31.9$ &$65.8$ & $43.3$   \\
M5  &$46.7$ &   $47.7$ &$56.3$ &$32.4$ &$66.2$ & $43.9$   \\
M6  &$47.4$ & $48.5$ & $\mathbf{56.5}$ &$33.7$  &$\mathbf{66.4}$ &$45.8$    \\
M7 (\net)  &$\mathbf{48.7}$ &   $\mathbf{50.4}$ &$\mathbf{56.5}$ &$\mathbf{37.1}$ & $\mathbf{66.4}$ &$\mathbf{51.6}$    \\
\bottomrule
\end{tabular}
\vspace{-0.4cm}
\end{table}

In this section, we quantitatively evaluate the influence of each proposed architectural component in PAPS, on the overall performance. Here, the addition of modules to the architecture of the base model M1 in the incremental form is performed according to their description in~\secref{sec:method}. \tabref{tab:instanceHeadEvaluation} presents results from this experiment. We begin with the model M1 which employs a semantic decoder as described in~\secref{subsubsec:semantic} without any cross-task module and occlusion segmentation head \myworries{and is similar to~\cite{cheng2020panoptic} with amodal capabilities}. For the instance decoder, it employs the aforementioned semantic decoder with the heads described in~\secref{subsubsec:instance} without occlusion-awareness of center and~\textit{thing} semantic segmentation. In the M2 model, we replace the instance decoder architecture with that described in~\secref{subsubsec:instance} without the cross-task module and the same heads as the M1 model. The improvement in performance shows the importance of multi-scale features from cross-layers for amodal instance center regression. In the M3 model, we add the \textit{thing} segmentation head to the instance decoder whose output is used during inference as described in~\secref{subsubsec:inference}. The improvement achieved indicates that the two decoders capture diverse representations of \textit{thing} classes which further improves the performance. 

In the M4 model, we add the context extractor module. The higher increase in APQ$_T$ compared to APC$_T$ indicates that the multi-scale features obtained from the aggregation of within-scales and cross-scales layers are much richer in the representation capacity, thereby improving the detection of small far away objects. Building upon M4, in the M5 model, we add the cross-task module. The increase in both \textit{stuff} and \textit{thing} components of the metrics demonstrates that the two decoders learn complementary features which when propagated bidirectionally is mutually beneficial for each of them. In the M6 model, we add the occlusion segmentation head and occlusion awareness to the inmodal center prediction head. We observe an improvement of $1.3\%\text{-}1.9\%$ in \textit{thing} components of the metrics demonstrating that the incorporation of occlusion information is integral for good amodal mask segmentation. Lastly, in the M7 model, we add the amodal mask refiner. The substantial improvement of $3.4\%$ and $5.8\%$ in APQ$_T$ and APC$_T$, respectively, demonstrates the efficacy of our proposed module. We note that the improvement in APC$_T$ is higher than APQ$_T$ indicating that the increase in segmentation quality of objects with larger occlusion areas is relatively higher than the smaller areas. This result precisely demonstrates the utility of our proposed amodal mask refiner, validating our idea of using embeddings of non-occluded object masks to supplement the amodal features with correlation for mid-to-heavy occlusion cases.\looseness=-1

\begin{table}
\footnotesize 
\centering
\caption{Evaluation of various propsal-based amodal panoptic segmentation approaches with our proposed amodal mask refiner. The performance is shown for the models trained on the BDD100K-APS dataset and evaluated on the validation set. Subscript $T$ refer to \textit{thing} classes. All scores are in [\%].}
\label{tab:amrEvaluation}
\begin{tabular}{p{1.5cm}c|p{0.6cm}p{0.6cm}|p{0.6cm}p{0.6cm}}
\toprule
Model & Amodal Mask Refiner & APQ & APC &  APQ$_T$ & APC$_T$\\
\midrule
ORCNN~\cite{follmann2019learning} & &$44.9$ &  $46.2$ &$29.9$ &$41.4$   \\

BCNet~\cite{Ke_2021_CVPR} &  &$45.2$ &   $46.4$ &$30.7$ &$42.1$    \\

ASN~\cite{qi2019amodal}   &  & $45.5$ &   $46.6$ &$31.2$ &$42.7$  \\

APSNet~\cite{\APSref} &  &$46.3$ &   $47.3$ &$32.8$ &$44.5$    \\
\midrule
ORCNN~\cite{follmann2019learning} & \checkmark &$45.3$ &  $46.6$ &$30.9$ &$42.8$   \\
BCNet~\cite{Ke_2021_CVPR} & \checkmark &$46.3$ &   $47.8$ &$33.2$ &$46.4$   \\
ASN~\cite{qi2019amodal}   &\checkmark &$46.7$ &   $48.1$ &$34.4$ &$47.1$    \\
APSNet~\cite{\APSref} &\checkmark  & $\mathbf{47.5}$ & $\mathbf{48.9}$  & $\mathbf{35.9}$ & $\mathbf{49.2}$    \\
\bottomrule
\end{tabular}
\end{table}

\begin{table}
\footnotesize
\centering
\caption{Performance comparison of panoptic segmentation on the Cityscapes validation set.  $-$ denotes that the metric has not been reported for the corresponding method. All scores are in [\%].}
\begin{tabular}{p{3cm}|p{0.3cm}p{0.3cm}p{0.3cm}p{0.3cm}p{0.3cm}p{0.3cm}p{0.5cm}}
\toprule
Network & PQ & SQ & RQ &PQ\textsubscript{T} & PQ\textsubscript{S} & AP & mIoU  \\
\midrule
 Panoptic FPN~\cite{kirillov2019bpanoptic} &  $58.1$ & $-$ & $-$ & $52.0$ & $62.5$ & $33.0$ & $75.7$ \\
 UPSNet~\cite{xiong2019upsnet} &  $59.3$ & $79.7$  & $73.0$ & $54.6$ &  $62.7$  & $33.3$ & $75.2$ \\
 DeeperLab~\cite{yang2019deeperlab} &  $56.3$ & $-$ & $-$ & $-$ &  $-$ & $-$ & $-$ \\
 Seamless~\cite{porzi2019seamless} &  $60.3$ & $-$ & $-$ & $56.1$ &  $63.3$ & $33.6$ & $77.5$ \\
 SSAP~\cite{Gao_2019_ICCV} &  $61.1$ & $-$ & $-$ & $55.0$ & $-$ & $-$ & $-$ \\
 AdaptIS~\cite{sofiiuk2019adaptis} &  $62.0$ & $-$ & $-$ & $58.7$ &$64.4$ &  $36.3$ & $79.2$ \\
 Panoptic-DeepLab~\cite{cheng2020panoptic} &  $63.0$ & $-$ & $-$ & $-$ &  $-$ &  $35.3$ & $80.5$ \\
 EfficientPS~\cite{mohan2020efficientps} &  $63.9$ & $81.5$  & $77.1$ & $\mathbf{60.7}$ & $66.2$  &$\mathbf{38.3}$ & $79.3$ \\
\midrule
\net (ours) &  $\mathbf{64.3}$ & $\mathbf{82.1}$  & $\mathbf{77.3}$ & $60.1$ & $\mathbf{67.3}$ & ${37.2}$ & $\mathbf{80.8}$ \\
\bottomrule
\end{tabular}
\label{tab:baselineCityscapes}
\vspace{-0.3cm}
\end{table}

\subsubsection{Generalization of amodal mask refiner}
\label{subsubsec:amr}

In this section, we study the generalization ability of our proposed amodal mask refiner by incorporating it in existing proposal-based amodal panoptic segmentation approaches. To do so, we adapt the amodal mask refiner by removing all downsampling layers in the encoders and upsampling layers from its decoder, to make it compatible with proposal-based approaches. We add an occlusion classification branch in the amodal instance head of all the proposal-based methods similar to ASN~\cite{qi2019amodal} and add another identical amodal mask head. The output of the fourth layer of the amodal mask head of each method is considered as the amodal features input. For the non-occluded object features, we multiply the output of the occlusion classification branch with the output of the fourth layer of the inmodal mask head. We feed the amodal features and non-occluded object features to the amodal mask refiner, followed by concatenating its output with the amodal features. Subsequently, we feed these concatenated features to the newly added amodal mask head. To train the networks, we use the same two-stage procedure described in \secref{subsec:training} and the training protocol described in~\cite{\APSref}.\looseness=-1 

\tabref{tab:amrEvaluation} presents the results from this experiment. We observe a considerable improvement in the performance of all the proposal-based methods demonstrating the effectiveness and the ease of integration into existing architectures. Moreover, the improvement achieved for APSNet is higher than ORCNN indicating that the performance can vary depending on the quality of the inmodal and amodal feature representations in the network.

\subsubsection{Panoptic Segmentation Results on Cityscapes Dataset}
\label{subsubsec:city}

In this section, we evaluate the performance of our proposed \net for panoptic segmentation on the Cityscapes~\cite{cordts2016cityscapes} dataset. In the architecture, we remove the amodal mask refiner, occlusion segmentation, amodal center offset, relative occlusion order segmentation, and amodal center regression heads as they only contribute to obtaining the amodal masks. We train our network with a learning rate $lr=0.001$ for $90$K iterations using the Adam optimizer. We report the Panoptic Quality (PQ), Segmentation Quality (SQ) and Recognition Quality (RQ) metrics on the validation set of Cityscapes for single-scale evaluation in \tabref{tab:baselineCityscapes}. For the sake of completeness, we also report the Average Precision (AP), and the mean Intersection-over-Union (mIoU) scores. We observe that \net achieves the highest PQ score of $64.3\%$ which is $1.3\%$ and $0.4\%$ higher than the state-of-the-art Panoptic-DeepLab and EfficientPS, respectively. The improvement achieved over Panoptic-DeepLab demonstrates the efficacy of our proposed modules and architectural design choices.

\subsubsection{Performance on KINS Dataset}
\label{subsubsec:kins}

\begin{table}
\footnotesize 
\centering
\caption{Amodal instance segmentation results on the KINS dataset. All scores are in [\%].}
\begin{tabular}{p{3cm}|p{1.5cm}p{1.5cm}}
\toprule
Model &  Amodal$_{AP}$ & Inmodal$_{AP}$ \\
\midrule
ORCNN~\cite{follmann2019learning}  & $29.0$ &$26.4$  \\
VQ-VAE~\cite{jang2020learning}  & $31.5$ &$-$  \\
Shape Prior~\cite{yuting2021amodal}  & $32.1$ &$29.8$  \\
ASN~\cite{qi2019amodal}  & $32.2$ &$29.7$   \\
APSNet~\cite{\APSref} & $35.6$  & $32.7$ \\
\midrule
\net (Ours)  & $\mathbf{37.4}$  & $\mathbf{33.1}$ \\
\bottomrule
\end{tabular}
\label{tab:kins_dataset}
\vspace{-0.4cm}
\end{table}

We benchmark the performance of our proposed \net architecture on the KINS~\cite{qi2019amodal} amodal instance segmentation benchmark. This benchmark uses the Average Precision (AP) metric for evaluating both amodal and inmodal segmentation. We train our network with a learning rate $lr=0.001$ for $40$K iterations using the Adam optimizer. We use the same validation protocols as \cite{qi2019amodal}. \tabref{tab:kins_dataset} presents results in which our proposed \net outperforms the state-of-the-art APSNet by $1.8\%$ and $0.4\%$ for amodal AP and inmodal AP, respectively, establishing the new state-of-the-art on this benchmark. The large improvement in the Amodal\textsubscript{AP} compared to the Inmodal\textsubscript{AP} indicates refining amodal masks with unoccluded object embeddings is an effective strategy.


\begin{figure*}
\centering
\footnotesize
{\renewcommand{\arraystretch}{0.5}
\begin{tabular}{p{0.4cm}P{4.6cm}P{4.6cm}P{4.6cm}}
&  \raisebox{-0.4\height}{APSNet~\cite{\APSref}} &  \raisebox{-0.4\height}{\mbox{PAPS} (Ours)} & 
\raisebox{-0.4\height}{Improvement\textbackslash{Error Map}}\\
\\
\rot{(a)} 
& \raisebox{-0.4\height}{\includegraphics[width=\linewidth]{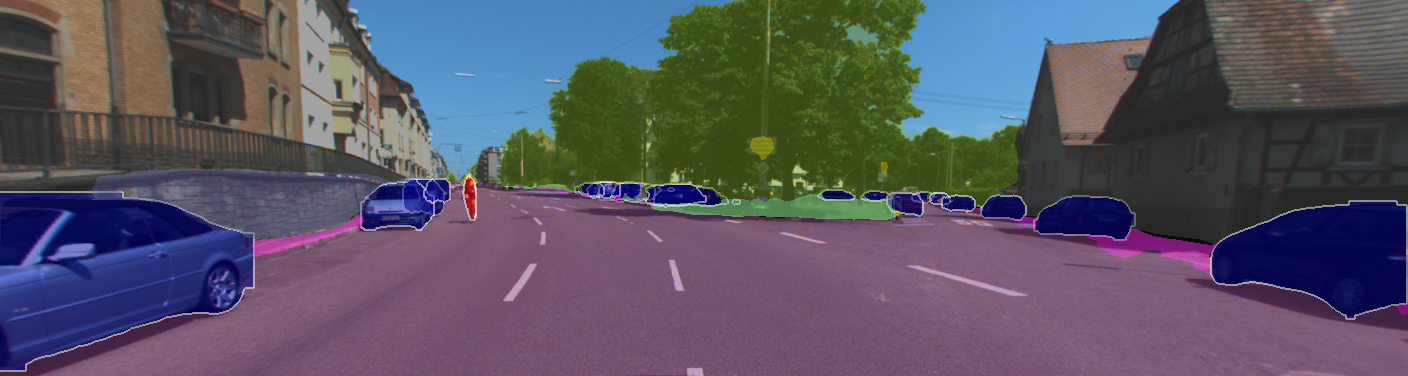}} & \raisebox{-0.4\height}{\includegraphics[width=\linewidth]{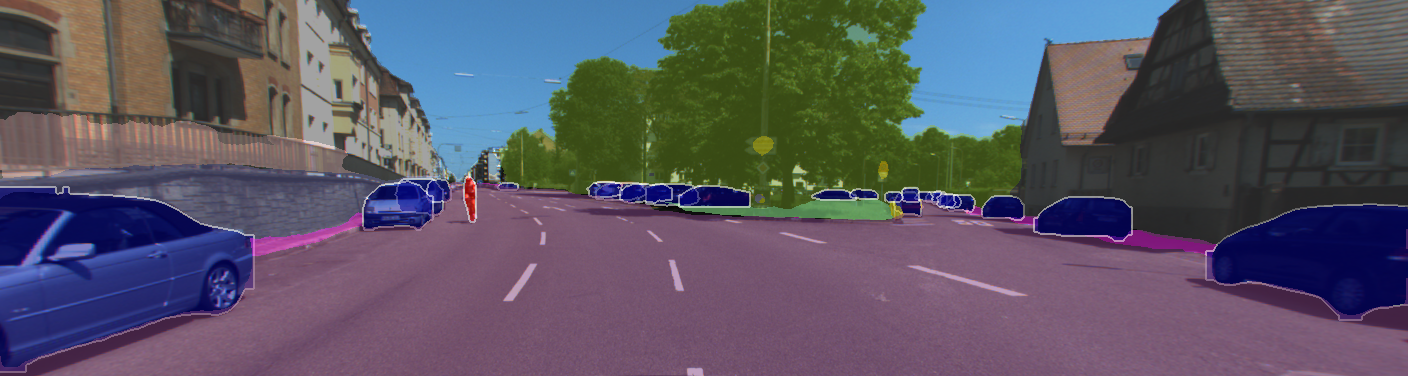}} & \raisebox{-0.4\height}{\includegraphics[width=\linewidth,frame]{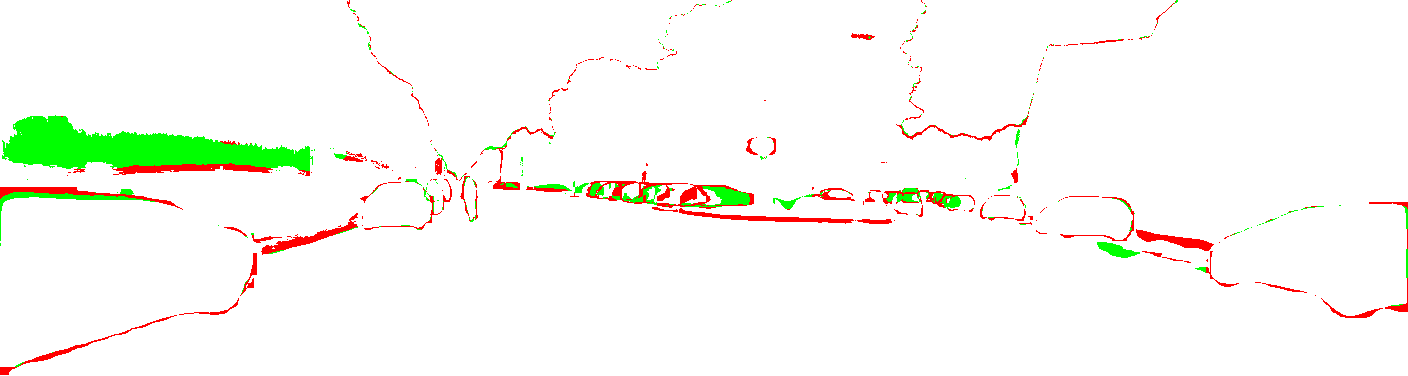}}\\
\\
\\
\rot{(c)} 
& \raisebox{-0.4\height}{\includegraphics[width=\linewidth]{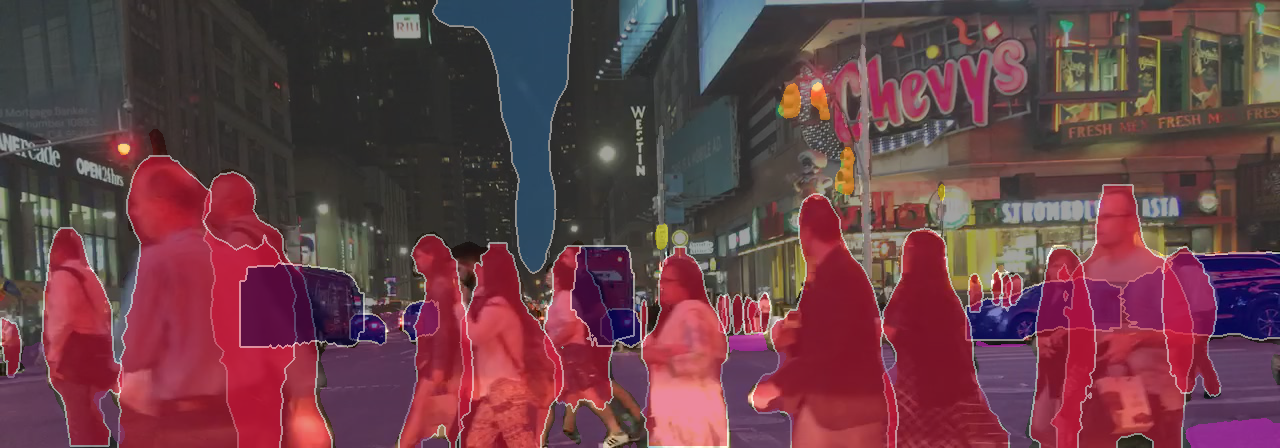}} & \raisebox{-0.4\height}{\includegraphics[width=\linewidth]{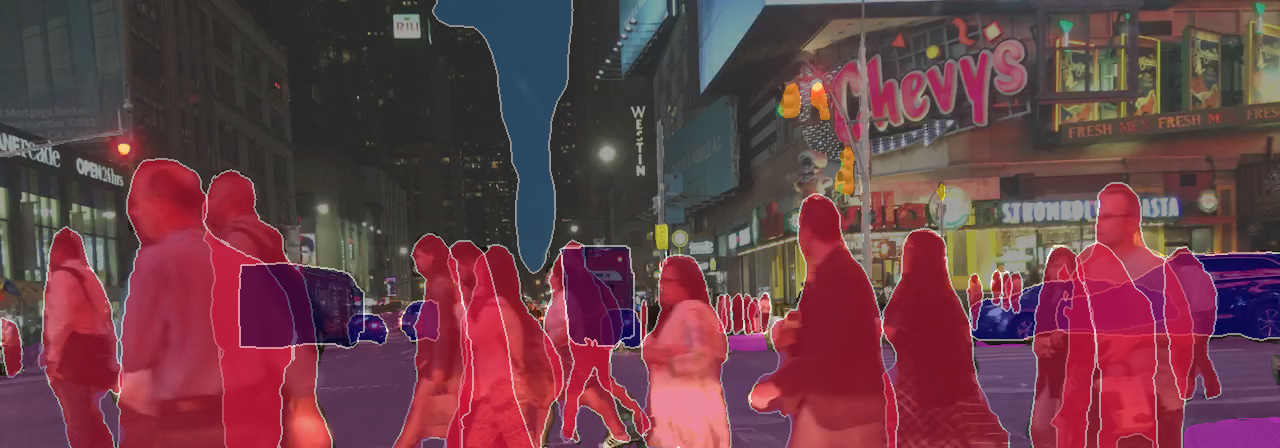}} & \raisebox{-0.4\height}{\includegraphics[width=\linewidth,frame]{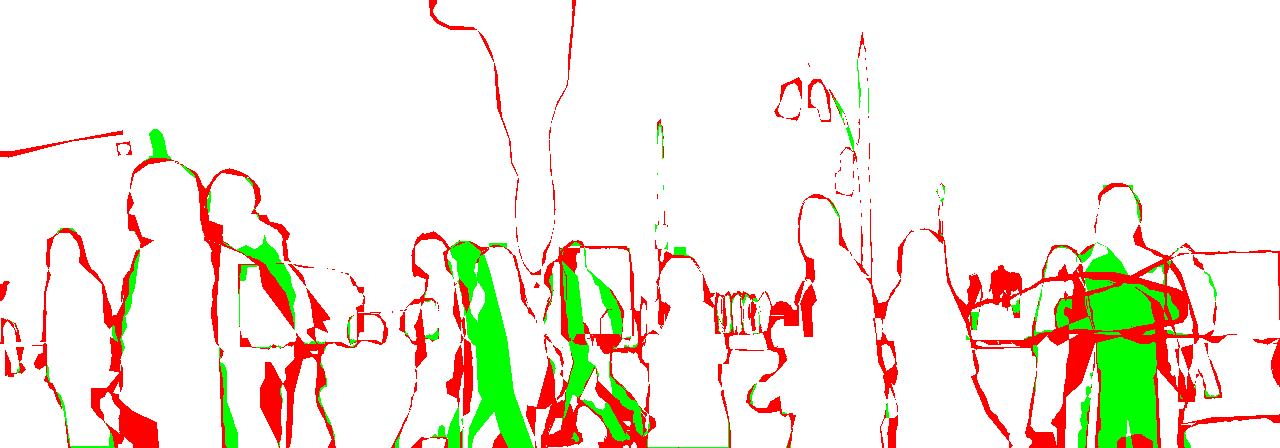}}\\
\\
\\
\end{tabular}}
\vspace{-0.3cm}
\caption{Qualitative amodal panoptic segmentation results of our proposed \mbox{PAPS} network in comparison to the state-of-the-art APSNet~\cite{\APSref} on (a) KITTI-360-APS and (b) BDD100K-APS datasets. We also show the Improvement\textbackslash{Error Map} which denotes the pixels that are misclassified by \mbox{PAPS} in red and the pixels that are misclassified by APSNet but correctly predicted by \mbox{PAPS} in green.} 
\label{fig:visual_ablation}
\vspace{-0.2cm}
\end{figure*}

\subsection{Qualitative Evaluations}
\label{subsec:qualitative}

In this section, we qualitatively compare the amodal panoptic segmentation performance of our proposed \mbox{PAPS} architecture with the previous state-of-the-art \mbox{APSNet}. \figref{fig:visual_ablation} presents the qualitative results. We observe that both approaches are capable of segmenting partial occlusion cases. However, our \mbox{PAPS} outperforms \mbox{APSNet} under moderate to heavy occlusion cases such as cluttered cars and pedestrians. In \figref{fig:visual_ablation}(a) the faraway cars on the right are detected more reliably by our network along with their amodal mask segmentations demonstrating the positive effects of within-scales and cross-scales multi-scale features and the occlusion aware heads. In \figref{fig:visual_ablation}(b), our model successfully predicts the amodal masks of heavily occluded pedestrians and cars. This demonstrates the utility of our amodal mask refiner module. By relying on the unoccluded mask features, PAPS is able to make a coarse estimate of the object's amodal masks. Furthermore, \mbox{PAPS} achieves more accurate segmentation of the challenging thin \textit{stuff} classes such as poles and fences.

\section{Conclusion}

In this work, we presented the first proposal-free amodal panoptic segmentation architecture that achieves state-of-the-art performance on both the KITTI-360-APS and BDD100K-APS datasets. To facilitate learning proposal-free amodal panoptic segmentation, our \net network learns amodal center offsets from the inmodal instance center predictions while decomposing the scene into different relative occlusion ordering layers such that there are no overlapping amodal instance masks within a layer. It further incorporates several novel network modules to capture within-layer multi-scale features for richer multi-scale representations, to enable bilateral propagation of complementary features between the decoders for their mutual benefit, and to integrate global and local occlusion features for effective amodal reasoning. Furthermore, we proposed the amodal mask refiner module that improves the amodal segmentation performance of occluded objects for both proposal-free and proposal-based architectures.  
Additionally, we presented detailed ablation studies and qualitative evaluations highlighting the improvements that we make to various core network modules of our amodal panoptic segmentation architectures. Finally, we have made the code and models publicly available to accelerate further research in this area.


\footnotesize
\bibliographystyle{IEEEtran}
\bibliography{references.bib}

\clearpage
\renewcommand{\baselinestretch}{1}
\setlength{\belowcaptionskip}{0pt}

\begin{strip}
\begin{center}
\vspace{-5ex}
\textbf{\LARGE \bf
Perceiving the Invisible: Proposal-Free Amodal Panoptic Segmentation} \\
\vspace{2ex}

\Large{\bf- Supplementary Material -}\\
\vspace{0.4cm}
\normalsize{Rohit Mohan and Abhinav Valada}
\end{center}
\end{strip}

\setcounter{section}{0}
\setcounter{equation}{0}
\setcounter{figure}{0}
\setcounter{table}{0}
\setcounter{page}{1}
\makeatletter

\renewcommand{\thesection}{S.\arabic{section}}
\renewcommand{\thesubsection}{S.\arabic{subsection}}
\renewcommand{\thetable}{S.\arabic{table}}
\renewcommand{\thefigure}{S.\arabic{figure}}


\let\thefootnote\relax\footnote{}%

\normalsize

In this supplementary material, we provide additional ablation studies on the proposed architectural components and the illustration of the context extractor module.  

\section{Ablation Study}
In this section, we first study the importance of the various components of our proposed cross-task module. Subsequently, we study the influence of the number of relative occlusion ordering layers on the performance of our network. For all the experiments, we train our PAPS network without the amodal mask refiner on the BDD100K-APS dataset and evaluate it on the validation set. We use APQ and APC metrics as the principal evaluation criteria for all the experiments performed in this section.

\subsection{Evaluation of the Cross-Task Module}
\label{subsubsec:crossexp}
In this section, we evaluate our proposed architecture of the cross-task module to enable bilateral propagation of features between the task-specific decoders. For this experiment, we use the PAPS architecture without the amodal mask refiner, similar to model~M6 in \secref{subsec:ablation}. \tabref{tab:crossHeadEvaluation} presents results from this experiment. We begin with model~M61 which does not use the cross-task module. In model~M62, we concatenate outputs of the opposite decoder as $F_O$. For the instance decoder, $F_O=F_S$ where $F_S$ are the output features of the semantic decoder. For the semantic decoder, $F_O=F_I$ where $F_I$ are the output features of the semantic decoder. The improvement in the performance shows the utility of propagating features between the task-specific decoders. In the model~M63, we define $F_O$ as the summation of the task-specific decoder features given as
\begin{align}
    F_{O} &= F_{I}+ F_{S}.
\end{align}

We observe a drop in performance for model~M63 compared to both model~M61 and model~M62 indicating that the use of summation fails to capture complementary features and at the same time affects learning the relevant primary features of the decoders themselves. In model~M64, we employ self-attention given by
\begin{align}
    F_{R} &= F_{I}+ F_{S},\\
    F_{O} &= g_3(F_{R})\cdot F_{R},
\end{align}
where $g_3(\cdot)$ is the function to compute the confidence scores of $F_R$. This model achieves improved performance over both model~M62 and model~M63 demonstrating that the attention mechanisms are beneficial for learning complementary features. As a next step, we employ self-attention to each individual task-specific decoder features in model~M65 and define $F_O$ as
\begin{align}
    F_{O} &= g_1(F_{I})\cdot F_{I} + g_2(F_{S})\cdot F_{S},
\end{align}
where $g_1(\cdot)$ and $g_2(\cdot)$ are the functions to compute the confidence scores. Model~M65 achieves a score lower than Model~M64 and similar to Model~M62. This indicates that applying self-attention to each input of the cross-task module effectively reduces them to be similar to a summation operation. Hence, in Model~M66, we employ cross-attention in $F_O$ as follows
\begin{align}
    F_{O} &= (1-g_1(F_{S}))\cdot F_{I} + (1-g_2(F_{I}))\cdot F_{S}.
\end{align}

This model achieves a performance similar to Model~M64 demonstrating that cross-attention is equally important as self-attention. Lastly, we use our proposed cross-attention followed by self-attention cross-task configuration (\eqref{eq:cross} and~\eqref{eq:cross2}), which yields the highest overall improvement. Consequently, from this experiment, we infer that cross-attention enables learning of adaptive complementary decoder features, whereas the following self-attention enables enhancement of these highly discriminative complementary features.

\begin{table}
\footnotesize 
\centering
\caption{Ablation study on various configurations of our proposed cross-task head. The performance is shown for the models trained on the BDD100K-APS dataset and evaluated on the validation set. Subscripts $S$ and $T$ refer to \textit{stuff} and \textit{thing} classes respectively. All scores are in [\%].}
\label{tab:crossHeadEvaluation}
\begin{tabular}{p{1.7cm}|p{0.6cm}p{0.6cm}|p{0.6cm}p{0.6cm}|p{0.6cm}p{0.6cm}}
\toprule
Model & APQ & APC &  APQ$_S$ &APQ$_T$ & APC$_S$ & APC$_T$\\
\midrule
M61  &$46.9$ &  $48.1$ &$56.1$ &$33.2$ & $66.0$ & $45.2$   \\
M62  &$47.0$ &   $48.1$ &$56.2$ &$33.3$ &$66.1$ & $45.3$   \\
M63  & $46.7$ &   $48.0$ &$55.9$ &$32.9$ &$65.9$ & $45.1$   \\
M64  &$47.1$ &   $48.2$ &$56.3$ &$33.4$ &$66.3$ & $45.4$   \\
M65  &$47.0$ &   $48.1$ &$56.2$ &$33.3$ &$66.1$ & $45.3$   \\
M66  &$47.1$ &   $48.2$ &$56.3$ &$33.4$ &$66.3$ & $45.4$   \\
M67  &$\mathbf{47.4}$ & $\mathbf{48.5}$ & $\mathbf{56.5}$ &$\mathbf{33.7}$  &$\mathbf{66.4}$ &$\mathbf{45.8}$    \\
\bottomrule
\end{tabular}
\end{table}

\subsection{Detailed Study on the Relative Occlusion Ordering Layers}
In this section, we study the effects of the number of relative occlusion ordering layers on the performance of our proposed architecture. Similar to \secref{subsubsec:crossexp}-A, for this experiment we use the PAPS architecture without the amodal mask refiner module. \tabref{tab:roo} shows results from this experiment. We begin with $N=4$ where \textit{N} is the number of relative occlusion ordering layers. The model achieves an improved score of $45.4\%$ and $46.6\%$ in APQ and APC, respectively compared to the baselines. This indicates that with four relative occlusion ordering layers, we can encapsulate sufficient object instances present in a given scene. Next, we use $N=6$ and obtain a significant improvement in the \textit{thing} components of the metrics. Subsequently, we train the model with $N=8$ which yields a lower performance in the metrics compared to $N=6$. This indicates that $N=6$ covers the majority of object instances in a given scene throughout the dataset. We then train the network with $N=10$ and $N=12$. These models do not achieve any improvement over the model with $N=8$ layers demonstrating that with eight relative occlusion ordering layers, we can encapsulate the maximal number of object instances in the dataset. 

\begin{table}
\footnotesize 
\centering
\caption{Influence on varying the number of layers of the relative occlusion ordering layers. The performance is shown for the models trained on the BDD100K-APS dataset and evaluated on the validation set. $N$ is the number of layers, subscripts $S$ and $T$ refer to \textit{stuff} and \textit{thing} classes respectively. All scores are in [\%].}
\label{tab:roo}
\begin{tabular}{p{1.7cm}|p{0.6cm}p{0.6cm}|p{0.6cm}p{0.6cm}|p{0.6cm}p{0.6cm}}
\toprule
N & APQ & APC &  APQ$_S$ &APQ$_T$ & APC$_S$ & APC$_T$\\
\midrule
4  &$45.4$ &  $46.6$ &$56.1$ &$29.3$ & $65.9$ & $41.1$   \\
6  &$46.8$ &   $47.8$ &$56.3$ &$32.6$ &$66.2$ & $44.3$   \\
8  &$\mathbf{47.4}$ & $\mathbf{48.5}$ & $\mathbf{56.5}$ &$\mathbf{33.7}$  &$\mathbf{66.4}$ &$\mathbf{45.8}$    \\
10  &$\mathbf{47.4}$ & $\mathbf{48.5}$ & $\mathbf{56.5}$ &$\mathbf{33.7}$  &$\mathbf{66.4}$ &$\mathbf{45.8}$    \\
12  &$\mathbf{47.4}$ & $\mathbf{48.5}$ & $\mathbf{56.5}$ &$\mathbf{33.7}$  &$\mathbf{66.4}$ &$\mathbf{45.8}$    \\
\bottomrule
\end{tabular}
\end{table}


\section{Context Extractor}
\begin{figure}
    \centering
    \begin{subfigure}[b]{0.7\linewidth}
        \centering
        \includegraphics[width=\textwidth]{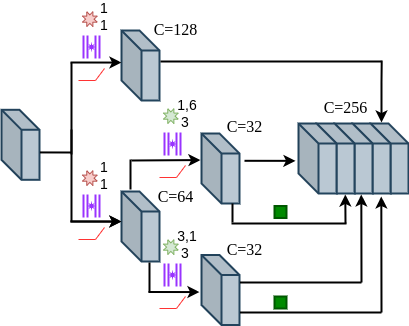}
        \label{ce_arch}
    \end{subfigure}
    \caption{Topology of our proposed context extractor module.}
    \label{fig:ce}
\end{figure}

Our proposed context extractor module enriches cross-scale features with within-scale contextual features, resulting in a rich multi-scale representation. This yields an improvement in performance for the instance decoder of our PAPS architecture as shown in \secref{subsec:ablation}-B. \figref{fig:ce} illustrates the architecture of the context extractor module. It splits the input into two parallel branches and employs two $1\times1$ convolutions. One of the branches is further subdivided into two parallel branches. Here, each branch uses a $3\times3$ depth-wise atrous separable convolutions with a dilation rate of $(1,6)$ and $(3,1)$, respectively. These branches are again subdivided into two parallel branches each. In each of these two parallel branches,  one branch employs a global pooling layer. Finally, all the outputs of the remaining parallel branches are concatenated. Please note that each of the convolutions is followed by batch normalization and ReLU activation function.

\end{document}